\documentclass[11pt,letterpaper,twocolumn]{template/style}

\usepackage{graphicx}
\usepackage{float,epstopdf}
\usepackage{algorithm}
\usepackage{algpseudocode}
\usepackage{bbm}

\usepackage{microtype}

\usepackage[numbers]{natbib}
\setcitestyle{square}

\usepackage{subcaption}
\usepackage{booktabs}

\usepackage{amsmath}
\usepackage{amssymb}
\usepackage{mathtools}
\usepackage{amsthm}
\usepackage{dsfont}
\usepackage{makecell}
\usepackage{multirow} 
\usepackage{amsfonts} 
\usepackage{mathrsfs}
\usepackage[amssymb, thickqspace]{SIunits}
\DeclareFontFamily{U}{rsfs}{\skewchar\font127}
\DeclareFontShape{U}{rsfs}{m}{n}{%
  <5> <5.5> <6> rsfs5
  <7> rsfs7
  <8> <9> <10> <10.95> <12> <14.4> <17.28> <20.74> <24.88> rsfs10
}{}
\DeclareFontShape{OML}{eur}{m}{n}{%
  <5> <5.5> <6> <7> <8> <9> gen * eurm
  <10> <10.95> <12> <14.4> <17.28> <20.74> <24.88> eurm10
}{}
\usepackage{enumitem}
\usepackage{pgfplotstable}
\pgfplotsset{compat=1.18}
\usepackage{lipsum}		

\usepackage{microtype}
\usepackage{graphicx}
\usepackage{booktabs} 
\usepackage[table]{xcolor}
\usepackage{arydshln}
\usepackage[normalem]{ulem} 

\usepackage{cases}
\usepackage{wrapfig}

\usepackage{url}

\usepackage{thmtools}
\usepackage{thm-restate}
\usepackage{tabu}

\definecolor{huskypurple}{HTML}{4B2E83}

\usepackage{titletoc}

\usepackage{listings}
\lstdefinestyle{promptstyle}{
  basicstyle=\ttfamily\footnotesize,
  breaklines=true,
  breakautoindent=false,
  breakindent=0pt,
  postbreak=\mbox{\textcolor{gray}{$\hookrightarrow$}\space},
  columns=fullflexible,
  keepspaces=true,
  frame=single,
  framesep=5pt,
  xleftmargin=6pt,
  xrightmargin=6pt,
  aboveskip=8pt,
  belowskip=8pt,
  showstringspaces=false,
}

\usepackage{fontawesome5}   

\makeatletter
\def\munderbar#1{\underline{\sbox\tw@{$#1$}\dp\tw@\z@\box\tw@}}
\makeatother

\AddToHook{cmd/appendix/before}{%
  \setcounter{axiom}{0}%
}

\newcommand{\be}{\begin{equation}}
\newcommand{\ee}{\end{equation}}
\newcommand{\bea}{\begin{equation*}\begin{aligned}}
\newcommand{\eea}{\end{aligned}\end{equation*}}

\newcommand{\monobold}[1]{{\fontfamily{lmtt}\bfseries\selectfont #1}}

\usepackage{pifont}

\title{ChebBooster: A Training-Free Approach for Efficient Diffusion Transformer Inference via Chebyshev-Inspired Extrapolation}
\runningtitle{ChebBooster: Chebyshev-Inspired Extrapolation for Efficient DiT Inference}

\newcommand{\affmark}[1]{%
  \textsuperscript{%
    {\usefont{T1}{pbk}{m}{n}\textcolor{BgPrimary600}{\textbf{#1}}}%
  }%
}

\author{Chengjie Lu\affmark{1}, Tianchi Deng\affmark{2}, Zhengqi He\affmark{1}, Chengwen Luo\affmark{2}, Xueliang Li\affmark{2}\affmark{†}\\
\affmark{1}College of Electronics and Information Engineering, Shenzhen University\\
\affmark{2}School of Artificial Intelligence, Shenzhen University\\
{\footnotesize\affmark{†}Corresponding author}\\
\faGithub~\textbf{Source Code:} \href{https://github.com/Kiramei/ChebBooster}{\monobold{https://github.com/Kiramei/ChebBooster}}
}

\begin{document}

\begin{abstract}

\vspace{-1mm}
\section*{Abstract}
Diffusion Transformers (DiTs) have shown strong performance in high-fidelity image generation, but their sampling process remains computationally intensive due to full model execution at every timestep. While cache-based acceleration has been explored to mitigate inference cost, naïve reuse schemes suffer from low accuracy over long intervals, and Taylor-series-based extrapolation methods often face instability caused by Runge oscillations. In this paper, we propose \textbf{ChebBooster}, a training-free extrapolation framework based on Chebyshev polynomial theory that achieves stable and efficient acceleration for DiTs. Specifically, we adopt the Barycentric formulation to evaluate Chebyshev approximants with high numerical stability and minimal overhead, and further decouple the extrapolation into an offline weight precomputation phase and a lightweight online application stage. Extensive experiments across three representative DiT-based models---DiT-XL/2, PixArt-\(\Sigma\), and FLUX.1-dev---demonstrate that ChebBooster achieves consistent improvements in visual quality and inference efficiency, reaching up to \textit{3.68$\times$} latency speedup and \textit{5.12$\times$} FLOPs reduction, outperforming existing training-free baselines under diverse generation tasks and resolutions.

\end{abstract}

\maketitle

\section{Introduction}

Denoising diffusion probabilistic models (DDPMs)~\cite{hoDDPM} have emerged as a dominant paradigm in generative modeling, surpassing traditional generative adversarial networks~\cite{Goodfellowgan, brock2019biggan, park2019SPADE} in synthesizing high-quality images from large-scale datasets. Recent advances have increasingly integrated Transformer architectures into diffusion models to better capture long-range dependencies, leading to the development of Diffusion Transformers (DiTs). DiTs have achieved remarkable performance across tasks such as class-to-image (C2I)~\cite{peebles2023dit} and text-to-image (T2I)~\cite{chen2023pixartalpha, chen2024pixartsigma, flux2024}. However, their growing capacity introduces a trade-off between generation quality and computational cost~\cite{liang2024scalinglawsdiffusiontransformers}.

\begin{figure}[ht]
    \centering
    \includegraphics[width=0.75\linewidth]{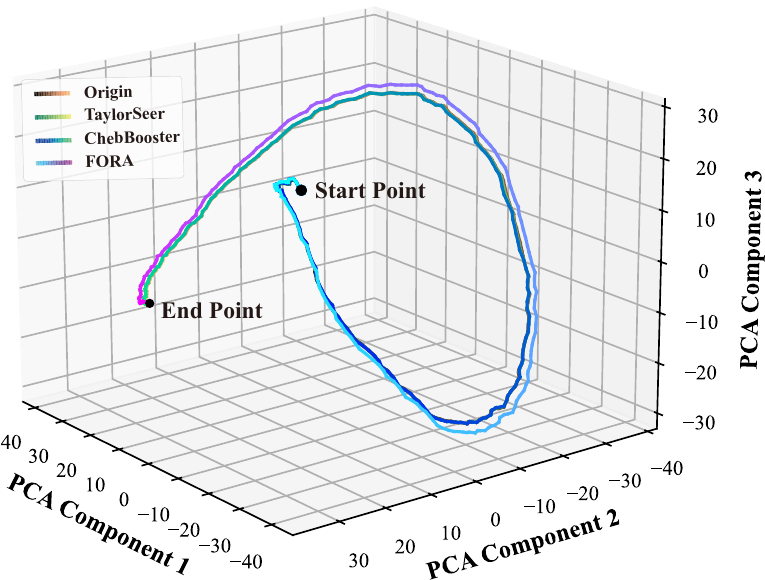}
    \caption{\textbf{3D Feature visualization of DiT forward features across timesteps.} The naïve caching method (i.e., FORA~\cite{selvaraju2024fora}) fails to align with the original feature trajectory, resulting in degraded generation quality.}
    \label{fig:3d_decomposition}
\end{figure}

To mitigate this trade-off, various acceleration strategies have been proposed. Quantization~\cite{li2023qdiffusion} and VAE-based training optimizations~\cite{yao2025vavae} reduce memory and training overhead, though they often require retraining or finetuning. Sampling-based approaches improve efficiency by reducing iteration counts through deterministic trajectories~\cite{songDDIM} or high-order solvers~\cite{lu2022dpm}, while flow-based models~\cite{liu2022flow, lauRealNVP, kingma2018glow} enable exact likelihood estimation but struggle with memory efficiency~\cite{helminger2021lossy}. Meanwhile, A complementary line of work focuses on inference-time caching~\cite{ma2024deepcache,selvaraju2024fora,toca,chen2024delta-dit,TaylorSeer2025}, where TaylorSeer~\cite{TaylorSeer2025} outperforms other methods with its mechanism transformation from cache-then-use to cache-then-forcast.

\begin{figure}[t]
    \centering
    \includegraphics[width=\linewidth]{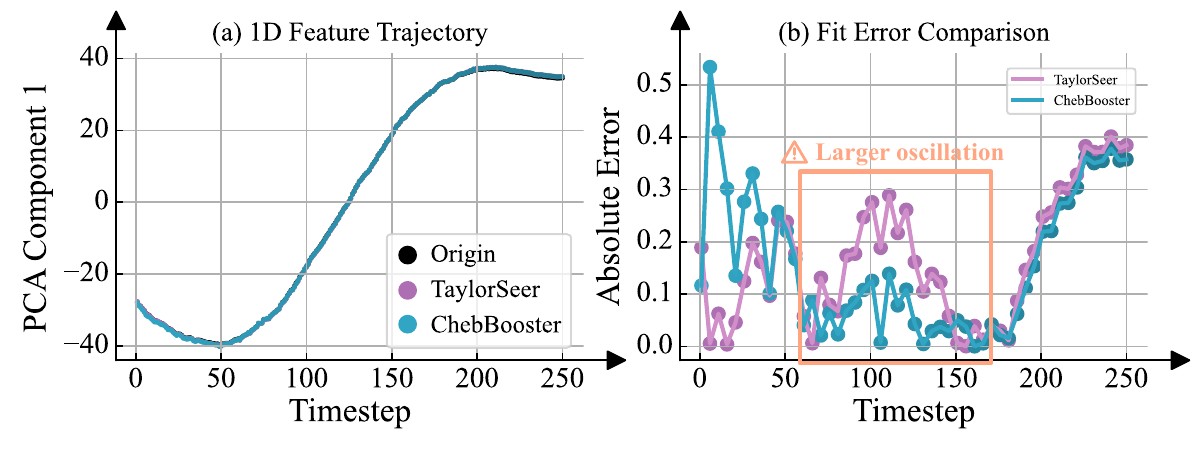}
    \caption{\textbf{1D feature comparison between TaylorSeer~\cite{TaylorSeer2025} and ChebBooster extrapolation.} (a) illustrates the feature trajectory of DiT, which evolves smoothly; both methods exhibit minimal differences in their fitted curves. (b) presents the extrapolation errors, where TaylorSeer displays larger oscillations in the mid-trajectory, yielding deviation in the predicted outcome.}
    \label{fig:1d_trajectory_comp}
    \vskip -0.2in
\end{figure}

To improve inference efficiency in diffusion models, various caching strategies have been proposed to reuse intermediate features during the denoising process. DeepCache~\cite{ma2024deepcache} exploits temporal redundancy by reusing high-level features across adjacent steps, while TeaCache~\cite{teacache} utilizes timestep embeddings to guide efficient, retraining-free caching. $\Delta$-DiT~\cite{chen2024delta-dit} further introduces a stage-aware mechanism to selectively cache computations based on the distinct roles of front and rear DiT blocks. Token-level reuse is also explored in ToCa~\cite{toca}, whereas FORA~\cite{selvaraju2024fora} directly stores attention maps with limited correction capacity at large timestep intervals (see Figure~\ref{fig:3d_decomposition}), restricting its effectiveness in aggressive acceleration regimes. To address this, TaylorSeer~\cite{TaylorSeer2025} models feature evolution via Taylor series expansion, enabling derivative-based extrapolation of future representations. However, its local approximation nature can introduce numerical instability, such as Runge oscillations~\cite{GrasselliPelinovsky2008}, especially in long-range predictions (see Figure~\ref{fig:1d_trajectory_comp}), ultimately degrading generation quality.

To tackle these problems, we propose a novel method called \textbf{ChebBooster}, which leverages Chebyshev-inspired extrapolation to predict future features of diffusion transformers. Chebyshev-based interpolation is a well-established technique in numerical analysis, known for mitigating the Runge phenomenon by approximating functions using Chebyshev polynomials~\cite{Trefethen2019approximation}. In our approach, we extend this idea by transforming interpolation into extrapolation, enabling future feature prediction across diffusion timesteps. Unlike standard Chebyshev interpolation, we adopt the barycentric formulation~\cite{berrutBarycentric}, which expresses the Lagrange interpolant as a rational function with precomputed weights. This formulation significantly improves numerical stability and reduces the computational overhead of evaluating the interpolant~\cite{floydcheb}. Building upon this efficient form, we notice that weight computation is merely dependent on caching schedule, while independent of the feature values themselves. This finding allows us to decouple the extrapolation into two stages: \textbf{(1) Offline precomputation stage} for the barycentric weights, and \textbf{(2) Online application stage} during inference. The weight table can be stored and reused locally, enabling repeated inferences under the same caching schedule to bypass the precomputation step---particularly beneficial for large-batch inference. Moreover, our method avoids redundant multiply--accumulate (MAC) operations by requiring only a single weight multiplication per extrapolation. Qualitative and quantitive experiments on three mainstream pretrained DiT-based image generation models---DiT/XL-2~\cite{hoDDPM}, PixArt-\(\Sigma\)~\cite{chen2024pixartsigma}, and FLUX.1-dev~\cite{flux2024}---across resolutions of \(256\times256\), \(512\times512\), and \(1024\times1024\) demonstrate that ChebBooster preserves generation quality while achieving up to \(\mathit{3.68\times}\) speedup in latency and \(\mathit{5.12\times}\) reduction in FLOPs under diverse settings.

\section{Related Works}

\subsection{Diffusion Model}

\begin{figure*}
    \centering
    \includegraphics[width=\textwidth]{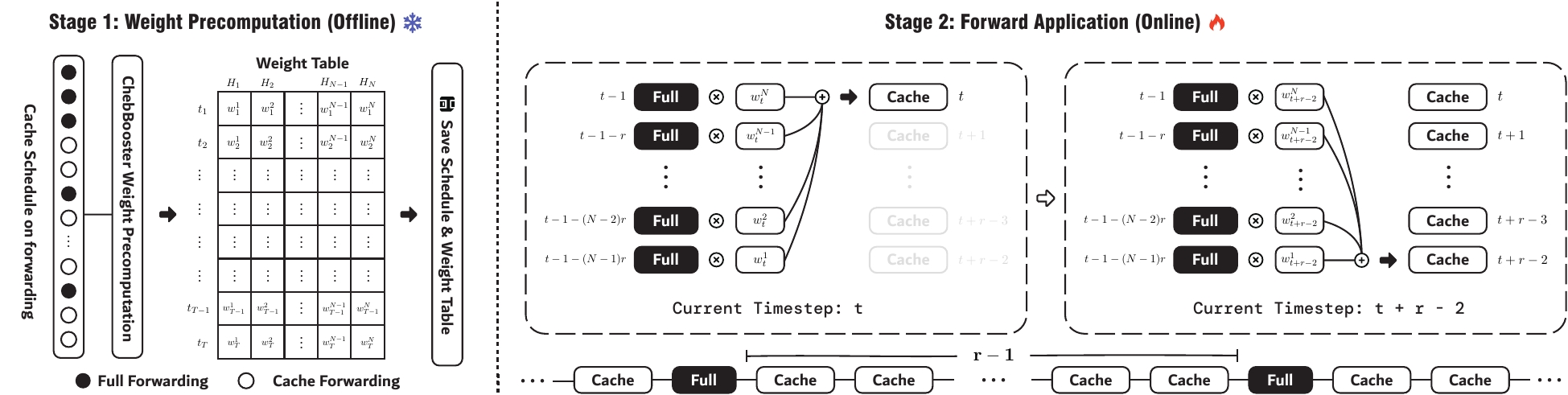}
    \caption{\textbf{Overview of the ChebBooster framework.} The process consists of two stages: (1) Weight Precomputation (offline), where Chebyshev-inspired extrapolation weights are precomputed for different timesteps and stored in a structured weight table guided by the cache schedule; and (2) Forward Application (online), where full model forwarding is performed at sparse reference timesteps, while intermediate features are efficiently extrapolated using cached activations and precomputed weights. This decoupled design enables acceleration by reducing redundant computations while maintaining generative fidelity.
    }
    \vskip -0.2in
    \label{fig:main_method_overview}
\end{figure*}

Diffusion models~\cite{hoDDPM} have become a cornerstone of generative modeling, excelling in high-quality image~\cite{peebles2023dit, chen2023pixartalpha, chen2024pixartsigma, flux2024} and video synthesis~\cite{kong2024hunyuanvideo, wan2025}. While early variants based on U-Net~\cite{hoDDPM, rombach2021highresolution} were effective for iterative denoising, they faced scalability challenges in high-resolution settings. To overcome these limitations, transformer-based architectures such as Diffusion Transformer (DiT)~\cite{peebles2023dit} have been introduced, leveraging the expressive power and scalability of transformers to enhance generation quality and flexibility. DiT has since evolved with techniques like self-conditioning~\cite{jiang2025sra, zhao2024dynamic, zhao2025dyditdynamicdiffusiontransformers} and novel normalization strategies~\cite{Zhu2025DyT}, extending its applicability to complex and multimodal tasks. Its scalability is further demonstrated by its success in synthesizing long, coherent video sequences~\cite{kong2024hunyuanvideo}, highlighting its promise in modeling real-world dynamics.

\subsection{Sampling Step Reduction}

Accelerating sampling while preserving output quality is a central goal in diffusion models. Original DDPMs~\cite{hoDDPM} require hundreds of steps, prompting step reduction methods such as DDIM~\cite{songDDIM}, which introduces a deterministic non-Markovian trajectory for faster sampling. The DPM-Solver series~\cite{lu2022dpm, lu2022dpm++, zheng2023dpm} further improves convergence via high-order ODE solvers. Alternative strategies include flow-based formulations like Rectified Flow~\cite{liu2022flow}, and knowledge distillation~\cite{salimans2022progressive, luhman2021knowledge}, which compress multi-step denoising into few-step student models. More recently, Consistency Models~\cite{song2023consistency} have shown promise, with extensions such as Generator-Augmented Consistency~\cite{issenhuth2024improving} and Truncated Consistency~\cite{lee2024truncated} enhancing few-step fidelity. Theoretical insights~\cite{yangimproved} favor two-step over single-step updates for stability, while continuous-time variants offer further scalability, solidifying consistency-based paradigms for efficient generation.

\subsection{Cache Acceleration}

Recent efforts to accelerate diffusion model inference have increasingly focused on caching intermediate features to exploit temporal redundancy between adjacent sampling steps. DeepCache~\cite{ma2024deepcache} first demonstrated that high-level U-Net features exhibit strong temporal consistency, enabling reuse via a simple, training-free non-uniform strategy with minimal quality loss. However, its reliance on U-Net limits generalizability to transformer-based architectures. FORA~\cite{selvaraju2024fora} extends caching to DiTs by reusing redundant attention and MLP features without modifying the model. AdaCache~\cite{kahatapitiya2024adaptive} further introduces adaptive scheduling by adjusting caching intervals based on feature residuals and motion dynamics. To handle non-uniform timestep changes, TeaCache~\cite{teacache} uses timestep embeddings and polynomial corrections to guide caching decisions. $\Delta$-DiT~\cite{chen2024delta-dit} proposes a structure-aware scheme that accelerates DiT blocks asymmetrically based on their roles in sampling. Token-level methods such as ToCa~\cite{toca} and TokenCache improve granularity by caching less salient tokens identified via attention or learned predictors, with layer-wise ratio adjustment to reduce error. Most recently, TaylorSeer~\cite{TaylorSeer2025} shifts from reuse to prediction, employing Taylor-series-based extrapolation to forecast future features, achieving high-fidelity generation under aggressive acceleration---without retraining.

\section{Method}

\subsection{Preliminary}
\label{sec:preliminary}
\textbf{Diffusion Models.} Diffusion models are a class of generative probabilistic models formulated through two stochastic processes: a forward process that incrementally perturbs data with noise, and a reverse process that learns to reconstruct data by denoising. Given a sample \( \mathbf{x}_0 \sim q(\mathbf{x}_0) \), the forward process produces a sequence \( \{\mathbf{x}_t\}_{t=1}^T \) over \( T \) timesteps as
\begin{equation}
\mathbf{x}_t =
\sqrt{\alpha_t} \, \mathbf{x}_{t-1} +
\sqrt{1-\alpha_t} \, \boldsymbol{\epsilon}_t,
\end{equation}
where \( \boldsymbol{\epsilon}_t \sim \mathcal{N}(\mathbf{0}, \mathbf{I}) \), and \( \{\alpha_t\} \) denotes a predefined noise schedule.

The reverse process approximates the intractable posterior \( q(\mathbf{x}_{t-1}|\mathbf{x}_t) \) via a parameterized Gaussian transition:
\begin{equation}
p_\theta(\mathbf{x}_{t-1}|\mathbf{x}_t) =
\mathcal{N}\big(
\mathbf{x}_{t-1};
\boldsymbol{\mu}_\theta(\mathbf{x}_t, t), \, \beta_t \mathbf{I}
\big),
\end{equation}
where the mean is computed by a neural network \( \boldsymbol{\epsilon}_\theta \) as
\begin{equation}
\boldsymbol{\mu}_\theta(\mathbf{x}_t, t) =
\frac{1}{\sqrt{\alpha_t}} \left(
\mathbf{x}_t -
\frac{1-\alpha_t}{\sqrt{1-\bar{\alpha}_t}} \,
\boldsymbol{\epsilon}_\theta(\mathbf{x}_t, t)
\right),
\end{equation}
with \( \bar{\alpha}_t = \prod_{i=1}^t \alpha_i \). The model parameters \( \theta \) are optimized to minimize a variational bound or an equivalent score-matching loss.

In the continuous-time setting, the diffusion process can be expressed as a stochastic differential equation:
\begin{equation}
d\mathbf{x} =
\sigma(t) \, d\mathbf{w},
\end{equation}
with the reverse dynamics governed by the score function \( \nabla_{\mathbf{x}} \log p_t(\mathbf{x}) \). This temporally-indexed generative framework forms the foundation for incorporating more expressive architectures and flexible inference algorithms.

\noindent \textbf{Blocks in Diffusion Transformers.}  
Building on the probabilistic foundation above, the DiT realizes the reverse process as a composition of \( L \) hierarchical blocks, expressed as \( G = B^1 \circ B^2 \circ \cdots \circ B^l \). Each block \( B^l \) (\( l=1, \dots, L \)) is modularly constructed from three components: a self-attention module \( S^l \), a cross-attention module \( C^l \) (when conditioning is required), and a feed-forward multilayer perceptron \( M^l \). For an input sequence of latent tokens \( x_t = \{ x_j \}_{j=1}^{H \cdot W} \) at timestep \( t \), where each \( x_j \) represents a patch embedding, the transformation performed by \( B^l \) is given by  
\begin{equation}
B^l(x) =
x + S^l(x) + C^l(x) + M^l(x),
\end{equation}  
where residual connections and adaptive normalization mechanisms are implicitly applied to maintain stable signal propagation.  
The modules \( S^l \), \( C^l \), and \( M^l \) adjust over timesteps to accommodate the changing noise profile inherent in the diffusion trajectory. This explicit modularization of \( G \) not only clarifies the architectural structure but also subtly suggests that each component might benefit from distinct fitting strategies.

\subsection{Chebyshev-Inspired Extrapolation}
\label{sec:cheb_extrapolation}
\begin{figure}[t]
    \centering
    \includegraphics[width=0.75\linewidth]{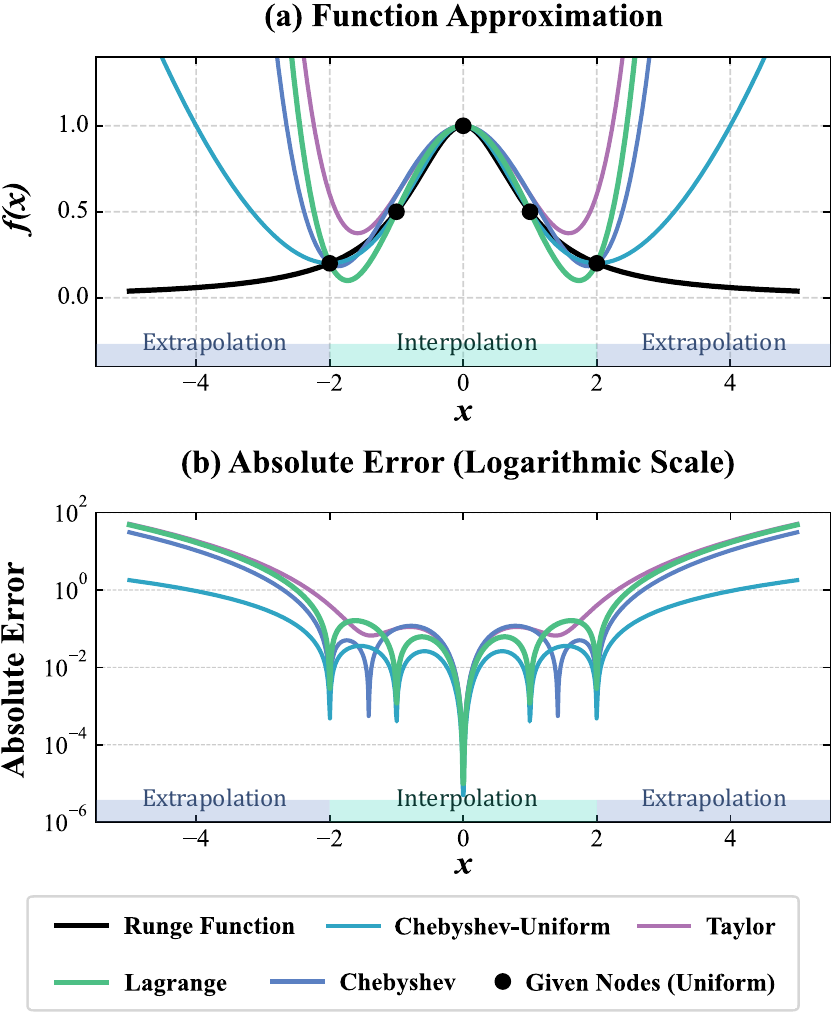}
    \caption{\textbf{Approximations and Errors with diverse methods on the Runge function.} Taylor expansion shows oscillation artifacts quickly in the near expolation. Lagrange interpolation starts solving the edge oscillation problem. Chebyshev methods give the best performance in both interpolation and extrapolation, while the one given equispaced nodes (Chebyshev-Uniform) is superior than that with the Chebyshev nodes.}
    \label{fig:method_selection}
    \vskip -0.2in
\end{figure}
The key in our method is to develop a training-free method for accelerating diffusion models by efficiently approximating future feature tensors, which requires a tool that can extrapolate a function's behavior from a small set of known points (i.e. cache schedule depicted in Figure~\ref{fig:main_method_overview}). We find that Chebyshev barycentric interpolation provides a robust and computationally ideal framework for this task. The foundation of this approach lies in polynomial interpolation. Given a function \( h(x) \) defined on \( x \in [-1, 1] \), we can approximate it with a polynomial. In the term of interpolation, using equispaced points for interpolation often leads to the Runge phenomenon, where oscillations near the interval's ends cause large errors. To overcome this, the Chebyshev method uses \textit{Chebyshev nodes}, the roots of the Chebyshev polynomials of the first kind, denoted by $T_N(x) = \cos(N \arccos x)$. For a polynomial of degree \(N\), these \(N\) nodes are given by:
\begin{equation}
    x_j = \cos \left( \frac{2j-1}{2N} \pi \right), \quad j=1, 2, \dots, N.
\end{equation}

\begin{table*}[t]
\centering

\setlength\tabcolsep{7pt} 

\resizebox{0.95\textwidth}{!}{
\begin{tabular}{l | c | c c c c | c c | c}
\toprule
\bf Method  & \makecell{\bf $\textit{r}$} & \bf Latency(s) $\downarrow$& \bf Speed $\uparrow$ & \bf FLOPs(T) $\downarrow$ & \bf Speed $\uparrow$  & \bf FID $\downarrow$ & \bf sFID $\downarrow$ &   \makecell{\bf Inception\\ \bf Score} $\uparrow$  \\
\midrule

{\textbf{$\text{DDIM-50 steps}$}}& \ding{56} & {0.506} & {1.000$\times$} & {23.735} & {1.000$\times$} & {2.18} & {4.29} & {251.56} \\
\midrule     

{\textbf{$\text{DDIM-25 steps}$}}& \ding{56} & {0.272} & {1.855$\times$} & {11.868} & {2.000$\times$} & {2.75} & {4.55} & {244.03} \\
{\textbf{$\text{TaylorSeer}$}}~\cite{TaylorSeer2025} & {\textbf{3}} & {0.343} & {1.473$\times$} & {8.570} & {2.770$\times$} & {2.34} & {4.76} & {247.00} \\
\rowcolor{gray!20}
{$\textbf{ChebBooster}$} ($H=3$) & {\textbf{3}}  & \textbf{0.302} & \textbf{1.674$\times$} & \textbf{8.564} & \textbf{2.771$\times$} & \textbf{2.25} & \textbf{4.59} & {246.49} \\
\rowcolor{gray!20}
{$\textbf{ChebBooster}$} ($H=5$) & {\textbf{3}}  & {0.342} & {1.476$\times$} & {8.567} & {2.770$\times$} & {2.28} & {4.64} & \textbf{247.27} \\
\midrule

{\textbf{$\text{DDIM-16 steps}$}}& \ding{56} & {0.189} & {2.681$\times$} & {7.595} & {3.125$\times$} & {4.24} & {5.69} & {223.38} \\
{\textbf{$\text{TaylorSeer}$}}~\cite{TaylorSeer2025} & {\textbf{4}} & {0.334} & {1.513$\times$} & {6.676} & {3.555$\times$} & {2.49} & {5.27} & \textbf{244.20} \\
\rowcolor{gray!20}
{$\textbf{ChebBooster}$} ($H=3$) & {\textbf{4}} & \textbf{0.279} & \textbf{1.812$\times$} & \textbf{6.668} & \textbf{3.560$\times$} & \textbf{2.35} & \textbf{4.95} & {242.58} \\
\rowcolor{gray!20}
{$\textbf{ChebBooster}$} ($H=5$) & {\textbf{4}} & {0.312} & {1.618$\times$} & {6.671} & {3.558$\times$} & {2.38} & {5.02} & \textbf{244.20} \\
\midrule 

{\textbf{$\text{DDIM-12 steps}$}}& \ding{56}  & {0.150} & {3.368$\times$} & {5.696} & {4.167$\times$} & {7.07} & {8.05} & {195.83} \\

{\textbf{$\text{TaylorSeer}$}}~\cite{TaylorSeer2025} & {\textbf{5}} & {0.304} & {1.662$\times$} & {5.725} & {4.146$\times$} & {2.63} & {5.46} & \textbf{241.83} \\
\rowcolor{gray!20}
{$\textbf{ChebBooster}$} ($H=3$) & {\textbf{5}} & \textbf{0.267} & \textbf{1.894$\times$} & \textbf{5.720} & \textbf{4.150$\times$} & \textbf{2.44} & \textbf{4.96} & {238.48} \\
\bottomrule
\end{tabular}}

\caption{\centering{\bf Quantitative comparison on DiT-XL/2 generation with resolution of 256$\times$256.}}
\label{table:cheb_s_result_table}
\end{table*}

\begin{table*}[t]
\centering

\setlength\tabcolsep{5.5pt}

\resizebox{0.95\textwidth}{!}{
\begin{tabular}{l | c | c c c c | c c}
    \toprule
    {\bf Method} & {\bf Refresh} &\multicolumn{4}{c|}{\bf Acceleration} &{\bf Image Reward $\uparrow$} &\bf CLIP$\uparrow$ \\
    \cline{3-6}
    
    {\bf FLUX.1~\cite{flux2024}} & {\bf Ratio ($\textit{r}$)} & {\bf Latency(s) $\downarrow$} & {\bf Speed $\uparrow$} & {\bf FLOPs(T) $\downarrow$}  & {\bf Speed $\uparrow$} & \bf DrawBench &\bf Score \\
    \midrule

{\textbf{$\text{[Dev]: 50 steps}$}} & \ding{56} & {26.039} & {1.000$\times$} & {3719.500} & {1.000$\times$} & {0.9613} & {31.63} \\
{\textbf{$\text{$\Delta$-DiT}$}}\cite{chen2024delta-dit} & {\textbf{2}} & {18.085} & {1.440$\times$} & {2480.000} & {1.500$\times$} & {0.9360} & {31.58} \\
\midrule

{\textbf{$\text{[Dev]: 30 steps}$}} & \ding{56} & {16.123} & {1.615$\times$} & {2231.700} & {1.667$\times$} & {0.9715} & {31.57} \\
{\textbf{$\text{$\Delta$-DiT}$}}\cite{chen2024delta-dit} & {\textbf{3}} & {13.381} & {1.946$\times$} & {1686.763} & {2.205$\times$} & {0.8969} & {31.53} \\
\midrule

{\textbf{$\text{[Dev]: 20 steps}$}} & \ding{56} & {13.470} & {1.933$\times$} & {1487.800} & {2.500$\times$} & {0.9838} & {31.42} \\
{\textbf{ToCa}~\cite{toca}} & {\textbf{5}} & {15.657} & {1.663$\times$} & {1064.060} & {3.496$\times$} & {0.9881} & {31.36} \\
{\textbf{$\text{TaylorSeer}$}~\cite{TaylorSeer2025}} & {\textbf{5}} & {8.256} & {3.154$\times$} & {893.730} & {4.162$\times$} & {0.9899} & {31.60} \\
\rowcolor{gray!20}
{$\textbf{ChebBooster}$ ($H=2$)} & {\textbf{5}} & \textbf{8.031} & \textbf{3.242$\times$} & \textbf{893.666} & \textbf{4.162$\times$} & \textbf{1.0070} & \textbf{31.63} \\
\midrule

{\textbf{$\text{[Dev]: 16 steps}$}} & \ding{56} & {8.952} & {2.909$\times$} & {1190.240} & {3.125$\times$} & {0.9281} & {31.13} \\
{\textbf{ToCa}~\cite{toca}} & {\textbf{6}} & {13.904} & {1.873$\times$} & {924.300} & {4.024$\times$} & {0.9771} & {31.25} \\
{\textbf{$\text{TaylorSeer}$}~\cite{TaylorSeer2025}} & {\textbf{6}} & {7.739} & {3.365$\times$} & {745.103} & {4.992$\times$} & {0.9946} & \textbf{31.69} \\
\rowcolor{gray!20}
{$\textbf{ChebBooster}$ ($H=2$)} & {\textbf{6}} & \textbf{7.078} & \textbf{3.679$\times$} & \textbf{744.938} & \textbf{4.993$\times$} & \textbf{0.9962} & \underline{31.61} \\
\bottomrule
\end{tabular}}
\caption{\centering \bf Quantitative comparison on FLUX.1[Dev] generation with resolution of 1024$\times$1024.}
\label{table:cheb_l_result_table}
\vskip -0.15in
\end{table*}

\begin{table*}[t]
\centering
\vskip -2mm

\setlength\tabcolsep{5.5pt}

\resizebox{0.95\textwidth}{!}{
\begin{tabular}{l | c | c c c c | c c}
    \toprule
    {\bf Method} & {\bf Refresh} &\multicolumn{4}{c|}{\bf Acceleration} &{\bf Image $\uparrow$} &\bf CLIP$\uparrow$ \\
    \cline{3-6}
    {\bf PixArt-$\mathbf{\Sigma}$~\cite{chen2024pixartsigma}} & {\bf Ratio ($\textit{r}$)} & {\bf Latency(s) $\downarrow$} & {\bf Speed $\uparrow$} & {\bf FLOPs(T) $\downarrow$}  & {\bf Speed $\uparrow$} & \bf Reward &\bf Score \\
    \midrule

{\textbf{$\text{DDIM-50 steps}$}} & \ding{56} & {1.255} & {1.000$\times$} & {95.362} & {1.000$\times$} & {1.1318} & {33.0278} \\

{\textbf{$\text{$\Delta$-DiT}$}~\cite{chen2024delta-dit}} & {\textbf{2}} & {0.469} & {2.675$\times$} & {36.138} & {2.639$\times$} & {1.1056} & {33.1066} \\

\midrule

{\textbf{$\text{DDIM-40 steps}$}} & \ding{56} & {1.003} & {1.250$\times$} & {76.290} & {1.250$\times$} & {1.1364} & {33.0198} \\
{\textbf{FORA}~\cite{selvaraju2024fora}} & {\textbf{3}} & {0.973} & {1.289$\times$} & {36.138} & {2.639$\times$} & {1.0917} & {33.0869} \\
{\textbf{ToCa}~\cite{toca}} & {\textbf{3}} & {1.265} & {0.992$\times$} & {58.238} & {1.637$\times$} & {1.0920} & {33.0869} \\
{\textbf{$\text{TaylorSeer}$}~\cite{TaylorSeer2025}} & {\textbf{3}} & {0.839} & {1.495$\times$} & {33.575} & {2.840$\times$} & {1.1195} & {33.0895} \\
\rowcolor{gray!20}
{$\textbf{ChebBooster}$ ($H = 3$)} & {\textbf{3}} & \textbf{0.706} & \textbf{1.778$\times$} & \textbf{33.540} & \textbf{2.843$\times$} & \textbf{1.1392} & \textbf{33.0943} \\
\midrule

{\textbf{$\text{DDIM-30 steps}$}} & \ding{56} & {0.753} & {1.666$\times$} & {57.217} & {1.667$\times$} & {1.1358}\textsuperscript{\textcolor{red}{†}} & {33.0194} \\
{\textbf{FORA}~\cite{selvaraju2024fora}} & {\textbf{4}} & {0.801} & {1.566$\times$} & {28.108} & {3.393$\times$} & {1.0609} & {33.0373} \\
{\textbf{ToCa}~\cite{toca}} & {\textbf{4}} & {1.176} & {1.067$\times$} & {53.312} & {1.789$\times$} & {1.0787} & {33.1463} \\
{\textbf{$\text{TaylorSeer}$}~\cite{TaylorSeer2025}} & {\textbf{4}} & {0.818} & {1.535$\times$} & {26.143} & {3.648$\times$} & {1.1111} & {33.1579} \\
\rowcolor{gray!20}
{$\textbf{ChebBooster}$ ($H = 4$)} & {\textbf{4}} & \textbf{0.622} & \textbf{2.018$\times$} & \textbf{26.112} & \textbf{3.652$\times$} & \textbf{1.1222} & \textbf{33.1699} \\
\midrule

{\textbf{$\text{DDIM-25 steps}$}} & \ding{56} & {0.622} & {2.018$\times$} & {47.681} & {2.000$\times$} & {1.1203}\textsuperscript{\textcolor{red}{†}} & {33.0418} \\
{\textbf{FORA}~\cite{selvaraju2024fora}} & {\textbf{5}} & {0.714} & {1.757$\times$} & {24.092} & {3.958$\times$} & {1.0023} & {32.9549} \\
{\textbf{ToCa}~\cite{toca}} & {\textbf{5}} & {1.128} & {1.112$\times$} & {50.687} & {1.881$\times$} & {1.0542} & {33.0372} \\
{\textbf{$\text{TaylorSeer}$}~\cite{TaylorSeer2025}} & {\textbf{5}} & {0.807} & {1.555$\times$} & {22.430} & {4.252$\times$} & {1.0660} & \textbf{33.2279} \\
\rowcolor{gray!20}
{$\textbf{ChebBooster}$ ($H = 2$)} & {\textbf{5}} & \textbf{0.536} & \textbf{2.339$\times$} & \textbf{22.362} & \textbf{4.265$\times$} & \textbf{1.1152} & \underline{33.0824} \\
\midrule

{\textbf{$\text{DDIM-20 steps}$}} & \ding{56} & {0.502} & {2.500$\times$} & {38.145} & {2.500$\times$} & {1.1090}\textsuperscript{\textcolor{red}{†}} & {33.0300} \\
{\textbf{FORA}~\cite{selvaraju2024fora}} & {\textbf{6}} & {0.631} & {1.990$\times$} & {19.073} & {5.000$\times$} & {0.9493} & {32.9773} \\
{\textbf{ToCa}~\cite{toca}} & {\textbf{6}} & {1.087} & {1.154$\times$} & {48.389} & {1.971$\times$} & {0.9873} & {33.1547} \\
{\textbf{$\text{TaylorSeer}$}~\cite{TaylorSeer2025}} & {\textbf{6}} & {0.794} & {1.580$\times$} & {18.710} & {5.097$\times$} & {1.0699} & {33.1878} \\
\rowcolor{gray!20}
{$\textbf{ChebBooster}$ ($H = 2$)} & {\textbf{6}} & \textbf{0.499} & \textbf{2.513}$\times$ & \textbf{18.640} & \textbf{5.116}$\times$ & \textbf{1.0784} & \textbf{33.1923} \\
\bottomrule
\end{tabular}}
\vspace{0mm}\
\footnotesize
\begin{itemize}\item \textsuperscript{\textcolor{red}{†}} Despite some performance retention, high FLOPs cost makes them unsuitable for efficient inference.\end{itemize}
\vspace{-2mm}
\caption{\centering \bf Quantitative comparison on PixArt-$\mathbf{\Sigma}$ generation with resolution of 512$\times$512.}
\label{table:cheb_m_result_table}
\vskip -0.15in

\end{table*}

These nodes are optimally distributed, clustering near the endpoints \( \pm 1 \), which guarantees stable and near-optimal polynomial approximation in theory. 

While one could construct the interpolating polynomial \( P_{N}(x) \) via a Chebyshev series expansion with Lagrange form, a more numerically stable and efficient representation is the \textbf{Barycentric interpolation formula}~\cite{berrutBarycentric}. This formula expresses the interpolant \( P_{N}(x) \) as a weighted average of the known function values \( h(x_j) \):
\begin{equation}
    P_{N}(x) = \displaystyle \sum_{j=1}^{N} \frac{\rho_j}{x - x_j} h(x_j) \Bigg / \displaystyle \sum_{j=1}^{N} \frac{\rho_j}{x - x_j},
\end{equation}
where the \( \rho_j \) are the precomputed barycentric weights. For Chebyshev nodes, these weights have a remarkably simple alternating form:
\begin{equation}
    \rho_j = (-1)^j \cdot \begin{cases} 0.5, & j=1 \text{ or } j=N, \\ 1, & \text{otherwise}. \end{cases}
\end{equation}
This formulation can be rewritten to highlight a crucial property for our application: the separation of weights and function values. Let us define the interpolation coefficients \( w^j_x \) as:
\begin{equation}
    \label{eq:barycentric_coeff}
    w^j_x = \frac{\rho_j}{x-x_j}\Bigg/\sum_{i=1}^{N} \frac{\rho_i}{x-x_i}.
\end{equation}
Then, the interpolation becomes a simple linear combination:
\begin{equation}
    P_{N}(x)  = \sum_{j=1}^{N} w^j_x h(x_j).
\end{equation}
This separability is the cornerstone of our method. The coefficients \( w^j_x \) depend only on the fixed node locations \( \{x_j\} \) and the evaluation point \( x \), but \emph{not} on the function values \( \{h(x_j)\} \). This allows us to pre-compute these coefficients for any target point. 

Finally, we intend to turn the interpolation problem into an extrapolation problem. To validate the problem more straightforward, we take the Runge Function as an example, which is defined as: \( f(x) = {(1 + 25x^2)}^{-1} \) .

As illustrated in Figure~\ref{fig:method_selection}, we investigate the behavior of various polynomial approximation strategies when applied to the function. Notably, our observations reveal that Chebyshev nodes, despite their theoretical advantages in interpolation, exhibit inferior performance compared to equispaced nodes in the context of extrapolation. Specifically, when approximating long-range target points beyond the original domain, extrapolation using Chebyshev nodes results in a significantly larger error scale. Based on this empirical evidence, we adopt equispaced nodes in our method to ensure more stable and accurate long-range prediction.

\subsection{ChebBooster}

Building upon the principles of Chebyshev-inspired extrapolation, we propose \textbf{ChebBooster}, a training-free acceleration scheme for DiTs, which is illustrated in Figure~\ref{fig:main_method_overview}. The key idea is to replace a subset of expensive full-network computations with lightweight extrapolation based on previously cached features.

Specifically, in this framework, we reinterpret Chebyshev polynomial extrapolation in the context of the diffusion process. Specifically, the function \( h(x) \) corresponds to a feature module's output tensor \( f \) within the diffusion model, while the variable \( x \) represents the diffusion timestep \( t \), normalized to a continuous variable \( \tau \in [-1, 1] \). The nodes \( \{x_j\} \) are the normalized timesteps \( \{\tau_j\} \) at which full computations are performed and the corresponding features \( \{f_j\} \) are cached. Given a query timestep \( t \) with normalized value \( \tau_t \), ChebBooster performs extrapolation to approximate the target feature \( f_t \) using the precomputed cached values. This formulation allows for accurate, efficient feature prediction across timesteps without retraining. ChebBooster operates in mainly two stages:

\noindent\textbf{1. Weight Precomputation.}
First, we define a schedule that dictates when to perform a full computation versus an extrapolation. Full computations occur at a set of timesteps \( s \in \mathcal{F} \), defined by:
\begin{equation}
    \mathcal{F} =
\bigl\{\, s \,\big|\, 
s \notin [s_0, T-1-s_1] 
\text{ or } 
r \mid (s-s_0) 
\bigr\}
\end{equation}
where \(s \in [0, T-1] \). The initial and final stages of diffusion reserve full computation, limited by \(s_0\) and \(s_1\), and \(r\) is the refresh ratio for intermediate steps. For all target timesteps \(t \notin \mathcal{F}\) where extrapolation will occur, we precompute the coefficients \( w^j_t \). This involves normalizing the target timestep \(t\) and the cached history timesteps \( \{s_j\} \) to the \( [-1,1] \) interval and applying Equation \ref{eq:barycentric_coeff}. These coefficients are computed once and stored.

\noindent\textbf{2. ChebBooster Forward Application.} During the denoising process, at each full-computation step \( s \in \mathcal{F} \), we compute the feature tensor \( f_s \), including \(S^l, C^l\) and \( M^l\). This tensor is detached from the computational graph and cached in a fixed-size history buffer \( H = \{(s_j, f_j)\}_{j=1}^{N} \), which holds the \( n \) most recent feature-step pairs. When the buffer is full, the oldest entry is discarded.

At any timestep \( t \notin \mathcal{F} \) where we wish to skip a full computation, and provided the history buffer \(H\) contains enough entries (\(|H|=n\)), we approximate the feature tensor \( f_t \) using the precomputed weights and the cached features:

\begin{equation}
    f_t = \sum_{j=1}^{N} w^j_t f_{s_j}, \quad \text{where } (s_j, f_{s_j}) \in H.
\end{equation}

This step is a simple, highly efficient linear combination of cached tensors, reducing the per-module complexity from that of a full network pass to just \(O(n)\). The entire set of schedules and weight tables can be prepackaged, making ChebBooster a portable and efficient drop-in accelerator for inference.

\section{Experiments}

\begin{figure*}[t]
    \centering
    \includegraphics[width=\linewidth]{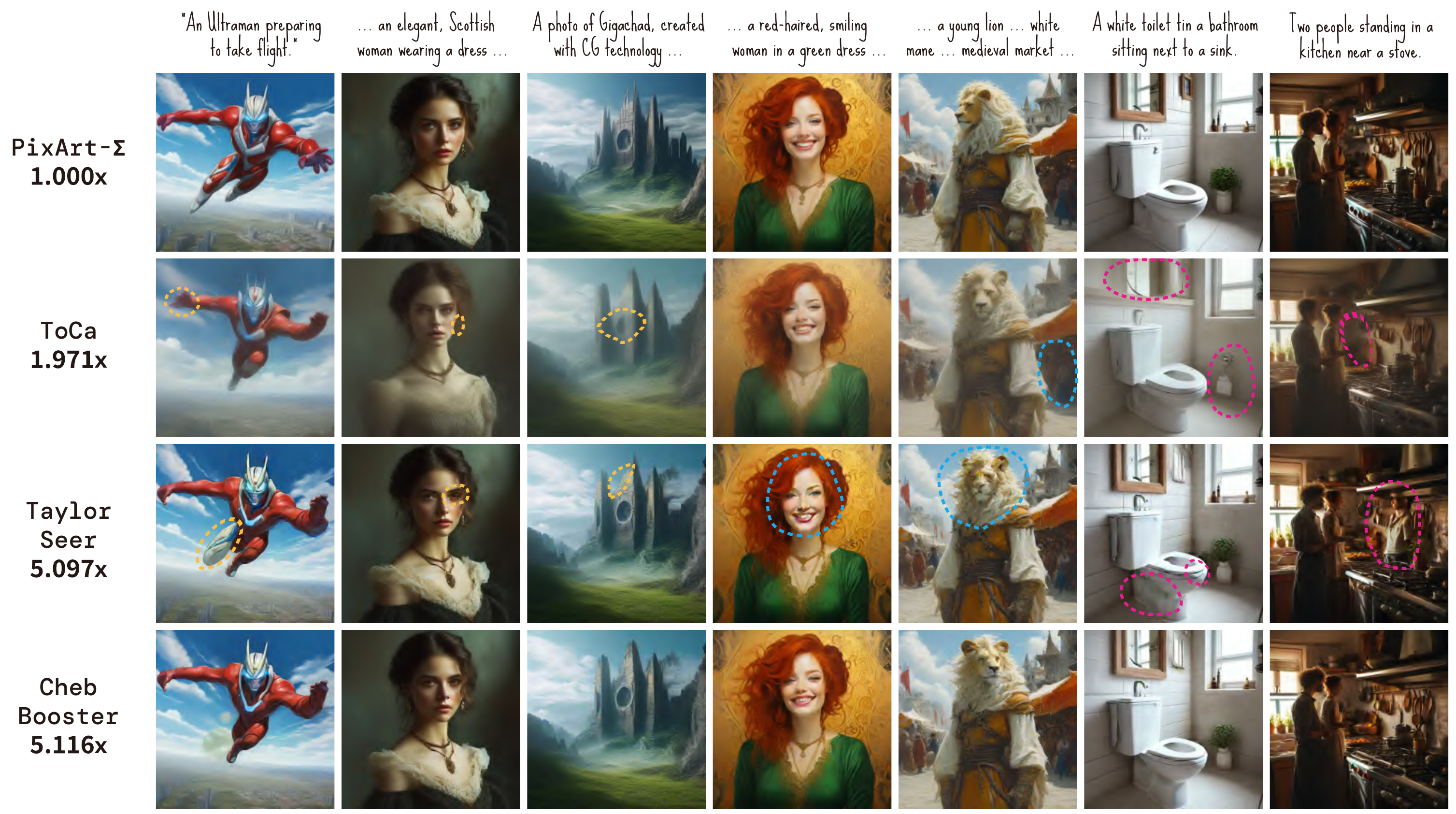}
    \caption{\centering{\bf Qualitative Study of diverse methods on PixArt-$\Sigma$.}}
    \label{fig:pixart_comparison}
    \vskip -0.15in
\end{figure*}

\subsection{Experimental Settings}

We evaluate our method on three DiT-based models: DiT-XL/2~\cite{peebles2023dit} (C2I, 256$\times$256), PixArt-\(\Sigma\)~\cite{chen2024pixartsigma} (T2I, 512$\times$512), and FLUX.1-dev~\cite{flux2024} (T2I, 1024$\times$1024), where quantitive results are displayed in Table~\ref{table:cheb_s_result_table}, \ref{table:cheb_m_result_table}, \ref{table:cheb_l_result_table}, respectively. All use official weights with 50 steps (DDIM for the first two, Rectified Flow for FLUX). DiT-XL/2 is trained on ImageNet~\cite{imagenet}; PixArt-\(\Sigma\) improves fidelity via weak-to-strong training and token compression; FLUX.1-dev enhances sharpness using self-attention and Rectified Flow. We generate 50K samples for DiT-XL/2 and evaluate them via FID~\cite{fid50k}, sFID~\cite{nashsfid}, IS~\cite{salimans2016improved}. We randomly choose 400 prompts from HPSv2~\cite{wu2023human} for PixArt-\(\Sigma\) and use all 200 DrawBench~\cite{saharia2022photorealistic} prompts for FLUX.1-dev, evaluated by CLIPScore~\cite{clipscore} and ImageReward~\cite{imageReward}, with the former measuring CLIP-based similarity and the latter modeling human preferences.

\subsection{Results on DiT-XL/2}
\label{subsec:dit-results}

We evaluate ChebBooster on DiT-XL/2 (256$\times$256) against state-of-the-art acceleration methods and reduced-step DDIM baselines. As shown in Table~\ref{table:cheb_s_result_table}, ChebBooster achieves consistently superior acceleration-quality trade-offs across various refresh ratios ($r$). At $r=3$, ChebBooster ($H=3$) yields the lowest FID of \textbf{2.25} with a \textbf{1.674$\times$} latency speedup --- \textbf{13.6\%} faster than TaylorSeer and \textbf{9.8\%} faster than DDIM-25. This corresponds to a \textbf{3.3\%} quality gain over DDIM-50 (FID = 2.18) with a \textbf{2.771$\times$} FLOPs reduction. At $r=4$, ChebBooster maintains \textbf{FID = 2.35} and \textbf{1.812$\times$} acceleration, outperforming TaylorSeer by \textbf{19.7\%} in speed and \textbf{5.6\%} in FID, and surpassing DDIM-20 (FID = 3.27) by \textbf{22.4\%} in latency. Under extreme acceleration ($r=5$), ChebBooster still delivers near-original quality \textbf{(FID = 2.44)} with \textbf{1.894$\times$} speedup, exceeding TaylorSeer by \textbf{14.0\%} in speed and \textbf{7.2\%} in FID, and significantly outperforming degraded DDIM-12.

\subsection{Results on PixArt-$\Sigma$}
\label{subsec:pixart-results}

As shown in Table~\ref{table:cheb_m_result_table}, ChebBooster achieves \textbf{state-of-the-art acceleration-quality trade-offs} on 512$\times$512 PixArt generation. At $r=3$ ($H=3$), it delivers \textbf{1.778$\times$} speedup---\textbf{19.0\%} faster than TaylorSeer---while attaining higher image quality (Reward=1.1392, $\Delta$+0.65\% vs. DDIM-50). At $r=4$ ($H=4$), ChebBooster reaches \textbf{2.018$\times$ speedup} with a \textbf{peak CLIP score of 33.1699}, exceeding TaylorSeer by \textbf{31.5\% in speed} and reducing FLOPs by \textbf{3.652$\times$}. The CLIP score also surpasses the DDIM-50 baseline by 0.43\%. Under aggressive settings ($r=6$, $H=2$), ChebBooster maintains \textbf{robust fidelity} (Reward=1.0784), while FORA and ToCa drop to 0.9493 ($\Delta$-16.0\%) and 0.9873 ($\Delta$-9.4\%), respectively.

For \textbf{Qualitative Study}, as shown in Figure~\ref{fig:pixart_comparison}, ChebBooster consistently outperforms ToCa and TaylorSeer, generating images with sharper structure, finer textures, and better semantic alignment to the prompts. For instance, in scenes like "... an elegant ... woman ...", ChebBooster preserves facial features and the eyes' details, while other methods exhibit noticeable distortions or omissions. These results highlight ChebBooster's robustness in maintaining high-fidelity generation even under aggressive acceleration.

\subsection{Results on FLUX.1-dev}
\label{subsec:flux-results}
\begin{figure}
    \centering
    \includegraphics[width=\linewidth]{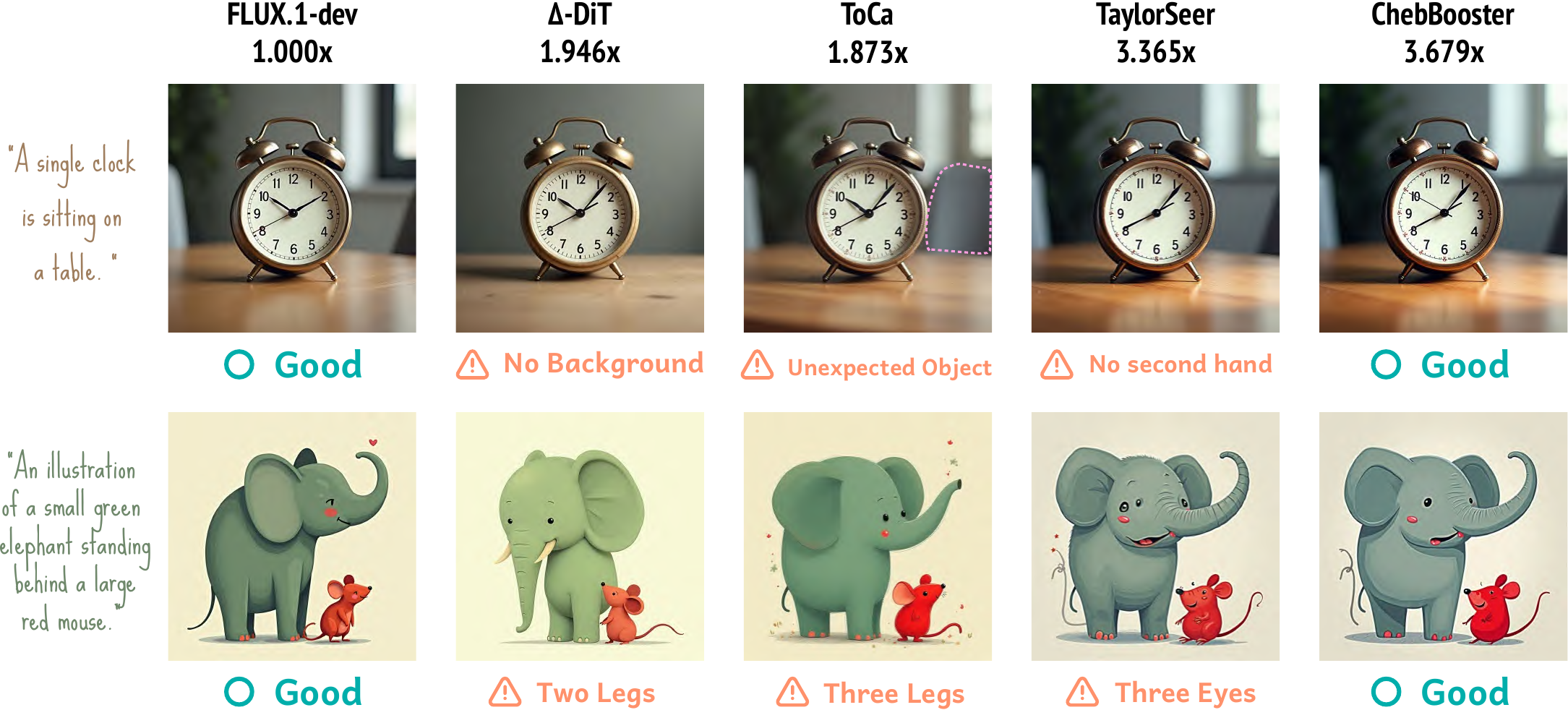}
    \caption{\bf \centering Qualitative Study on FLUX.1-dev.}
    \label{fig:comparison_result_flux}
\end{figure}

ChebBooster sets a new state-of-the-art for high-resolution generation on FLUX.1-dev, as shown in Table~\ref{table:cheb_l_result_table}. At $r=5$ ($H=2$), it achieves a \textbf{3.242$\times$} latency speedup (8.031s) with statistically superior image quality (Reward=1.0070, $\Delta$+4.76\% vs. DDIM-50), outperforming TaylorSeer and ToCa by \textbf{2.7\%} and \textbf{48.7\%} in speed, respectively, while reducing FLOPs by \textbf{4.162$\times$}. At $r=6$, ChebBooster reaches the fastest reported runtime of 7.078s (\textbf{3.679$\times$} speedup), 9.3\% faster than TaylorSeer, and matches baseline CLIP score (31.61 vs. 31.63). It also preserves Image Reward of \textbf{0.9962}, outperforming ToCa (0.9771, $p<0.01$) and FORA (0.9493) by \textbf{7.3\%} and \textbf{4.9\%}, under a \textbf{4.993$\times$} FLOPs reduction.

For \textbf{Qualitative Study}, as shown in Figure~\ref{fig:comparison_result_flux}, ChebBooster achieves the best visual consistency among all methods, accurately following prompt semantics while preserving object structure. ChebBooster avoids hallucinations like extra limbs or missing elements, which are present in other methods. These results confirm ChebBooster's robustness in high-resolution generation under strong acceleration.

\section{Conclusion}

We proposed ChebBooster, a training-free acceleration framework for Diffusion Transformers that leverages Chebyshev polynomial extrapolation to predict future features and reduce redundant computation during sampling. To address numerical instability and computational overhead traditionally associated with polynomial extrapolation, ChebBooster employs the Barycentric interpolation formulation for efficient and stable evaluation, and further decouples the extrapolation process into offline weight precomputation and lightweight online application. Experiments on multiple DiT-based models---covering C2I and T2I tasks at varying resolutions---demonstrate that ChebBooster achieves superior acceleration-quality trade-offs, outperforming prior cache-based methods in both visual fidelity and computational efficiency. With high acceleration ratio, ChebBooster provides a practical and scalable solution for generation and large-batch inference, and opens avenues for future exploration into adaptive caching and hybrid extrapolation strategies in generative modeling.

{\small
\raggedright
\bibliography{main}
\bibliographystyle{template/abbrvnat}
}

\clearpage
\appendix
\section{Extensive Proof for Chebyshev-Inspired Extrapolation}
This section establishes the mathematical basis for Barycentric Extrapolation with Equispaced Nodes. We show that, despite the traditional preference for Chebyshev nodes in interpolation, equispaced nodes---when paired with properly designed barycentric weights---offer a more stable and effective approach for extrapolation from sparse data. Unlike interpolation, where Chebyshev nodes minimize Runge's phenomenon, extrapolation benefits from the predictable behavior of the nodal polynomial in the equispaced setting, supporting our empirical findings.

Let us begin by establishing the general form of the barycentric interpolation formula, which is valid for any set of distinct nodes $\{x_j\}_{j=1}^N$. The foundation is the unique Lagrange interpolating polynomial $P_N(x)$ of degree at most $N-1$ that passes through the $N$ points $(x_j, h(x_j))$.

The Lagrange form is given by:
\begin{equation}
    P_N(x) = \sum_{j=1}^{N} h(x_j) l_j(x),
\end{equation}
where $l_j(x)$ are the Lagrange basis polynomials:
\begin{equation}
    l_j(x) = \prod_{\substack{k=1 \\ k \neq j}}^{N} \frac{x - x_k}{x_j - x_k}.
\end{equation}
Let us define the nodal polynomial $\omega(x)$ as:
\begin{equation}
    \omega(x) = \prod_{k=1}^{N} (x - x_k).
\end{equation}
The denominator of $l_j(x)$ can be expressed using the derivative of $\omega(x)$:
\begin{equation}
    \omega'(x_j) = \prod_{\substack{k=1 \\ k \neq j}}^{N} (x_j - x_k).
\end{equation}
Thus, the Lagrange basis polynomial can be rewritten as:
\begin{equation}
    l_j(x) = \frac{\omega(x)}{(x - x_j)\omega'(x_j)}.
\end{equation}
Substituting this back into the Lagrange formula yields the first barycentric form:
\begin{equation}
    P_N(x) = \omega(x) \sum_{j=1}^{N} \frac{\rho_j}{x - x_j} h(x_j),
\end{equation}
where the \textbf{barycentric weights}, $\rho_j$, are defined as:
\begin{equation} \label{eq:general_weights}
    \rho_j = \frac{1}{\omega'(x_j)}.
\end{equation}
This definition is universal and applies to any choice of distinct nodes. To derive the more numerically stable second form, we apply the same formula to the constant function $g(x) = 1$. Since $P_N(x)$ must be an exact interpolant, we have $1 = \sum_{j=1}^{N} l_j(x)$. This leads to:
\begin{equation}
    1 = \omega(x) \sum_{j=1}^{N} \frac{\rho_j}{x - x_j}.
\end{equation}
Dividing the polynomial formula by this identity, we cancel the $\omega(x)$ term and arrive at the final barycentric interpolation formula:
\begin{equation} \label{eq:barycentric_final}
    P_{N}(x) = \frac{\displaystyle \sum_{j=1}^{N} \frac{\rho_j}{x - x_j} h(x_j)}{\displaystyle \sum_{j=1}^{N} \frac{\rho_j}{x - x_j}}.
\end{equation}
This formulation is the cornerstone of the method. The key insight is that the weights $\rho_j$ are determined solely by the geometry of the nodes $\{x_j\}$, not the function values $\{h(x_j)\}$.

Here we make the justification for Equispaced Nodes in Extrapolation from two perspectives: the error in polynomial approximation and the growth of the nodal polynomial.

\noindent \textbf{The Error in Polynomial Approximation.}
The error of polynomial interpolation for a function $f(x) \in C^N([-1, 1])$ is given by the formula:

\begin{align}
    E(x) = f(x) - P_N(x)
    &= \frac{f^{(N)}(\xi)}{N!} \omega(x) \notag \\
    &= \frac{f^{(N)}(\xi)}{N!} \prod_{j=1}^{N} (x - x_j)
\end{align}

for some $\xi \in [-1, 1]$.

The primary goal of Chebyshev nodes is to minimize the term $\|\omega(x)\|_{\infty}$ for $x \in [-1, 1]$. The nodal polynomial for Chebyshev nodes, $\omega_C(x)$, is a scaled version of the Chebyshev polynomial $T_N(x)$, which has the unique property of having the smallest possible maximum magnitude on $[-1,1]$ among all monic polynomials of degree $N$. This guarantees near-optimal stability for \textit{interpolation}.

However, this guarantee does not apply to \textit{extrapolation}, i.e., when $|x| > 1$. Outside the interval $[-1, 1]$, the Chebyshev polynomial $T_N(x) = \cosh(N \cdot \text{arccosh}(x))$ grows exponentially fast. This rapid growth in $|\omega_C(x)|$ for $|x|>1$ can lead to a large extrapolation error, as observed empirically.

\noindent \textbf{Nodal Polynomial Growth for Equispaced Nodes.}
For equispaced nodes on $[-1, 1]$, given by $x_j = -1 + 2\frac{j-1}{N-1}$ for $j=1, \dots, N$, the nodal polynomial $\omega_E(x)$ does not possess the equi-oscillation property inside the interval. Its magnitude grows significantly towards the endpoints, leading to the Runge phenomenon for high $N$.

However, for extrapolation ($|x| > 1$), the growth of $|\omega_E(x)|$ can be more moderate compared to $|\omega_C(x)|$. The equispaced nodes are evenly distributed, preventing the multiplicative effect of $(x-x_j)$ from becoming disproportionately large, as it does for the clustered Chebyshev nodes when $x$ is far from the interval. In the context of sparse data, the number of nodes $N$ is inherently small. For small $N$, the Runge phenomenon is not a dominant factor, and the stability difference between node sets for in-domain interpolation is less critical. The primary concern becomes the behavior of the extrapolant, which is governed by the growth of $|\omega(x)|$ outside the domain. The choice of equispaced nodes is therefore a pragmatic and theoretically sound decision to control the growth of the extrapolation error.

Our method does not use "custom" weights in an ad-hoc manner; rather, it employs the mathematically correct barycentric weights corresponding to an equispaced grid, as derived from the general formula in Equation~\ref{eq:general_weights}.

Let the equispaced nodes be $x_j = -1 + (j-1)h$ for $j=1, \dots, N$, where the step size is $h = \frac{2}{N-1}$. The term $\omega'(x_j)$ is:

\begin{align}
    \omega'(x_j) 
    &= \prod_{k=1,\, k \neq j}^{N} (x_j - x_k) \\
    &= \prod_{k=1,\, k \neq j}^{N} \Big[  (-1 + (j{-}1)h) \notag \\ 
    &\qquad\quad - (-1 + (k{-}1)h) \Big] \\
    &= \prod_{k=1,\, k \neq j}^{N} (j-k)h \\
    &= h^{N-1} \prod_{k=1,\, k \neq j}^{N} (j-k) \\
    &= h^{N-1} \cdot (j-1)! \cdot (-1)^{N-j} (N-j)!
\end{align}

The barycentric weight $\rho_j$ is the reciprocal:
\begin{equation}
    \rho_j = \frac{1}{\omega'(x_j)} = \frac{1}{h^{N-1} (j-1)! (-1)^{N-j} (N-j)!}.
\end{equation}
We can absorb the constant scaling factor $h^{-(N-1)}$ into the overall normalization of Equation~\ref{eq:barycentric_final}, as it cancels out from the numerator and denominator. We can also adjust the alternating sign. A common convention is to define the weights as:
\begin{equation}
    \rho_j = (-1)^{j-1} \binom{N-1}{j-1}.
\end{equation}
This formulation is derived by recognizing that $\frac{(N-1)!}{(j-1)!(N-j)!} = \binom{N-1}{j-1}$ and adjusting the sign and constant factors. The use of these specific, alternating binomial coefficients as weights is therefore not an arbitrary choice but the direct consequence of applying the barycentric principle to an equispaced set of nodes.

In conclusion, the proposed method adopted in Chebyshev-inspired Expolation is a theoretically robust and well-justified technique. Its strength arises from a deliberate set of choices tailored to the problem of extrapolation from sparse data:
\begin{enumerate}
    \item \textbf{Barycentric Formulation:} It leverages the numerical stability and computational efficiency ($O(N)$ per evaluation point) of the barycentric formula (Equation~\ref{eq:barycentric_final}).
    \item \textbf{Equispaced Nodes:} It prioritizes stability in the extrapolation domain ($|x|>1$) over optimality in the interpolation domain ($|x| \le 1$). This is justified because the error term $|\omega(x)|$ for equispaced nodes exhibits more controlled growth outside the interval compared to the explosive growth associated with Chebyshev nodes, which is particularly relevant for long-range predictions.
    \item \textbf{Correct Weights:} It utilizes the mathematically derived barycentric weights for an equispaced grid, $\rho_j \propto (-1)^{j-1}\binom{N-1}{j-1}$, ensuring that the method correctly implements Lagrange polynomial extrapolation in a stable form.
\end{enumerate}
In summary, the method does not contradict established theory but rather applies it judiciously, recognizing that the optimal choice of nodes is context-dependent. For the specified goal of extrapolation from a limited number of points, the combination of equispaced nodes and their corresponding barycentric weights provides a superior and more reliable framework than the traditional Chebyshev-based approach.

\section{Pseudocode of ChebBooster}

From provided pseudocode in Algorithm~\ref{alg:chebbooster_pseudocode}, we can make it clearer that our ChebBooster algorithm consists of two main stages: an offline precomputation stage and an online application stage, which makes the cache and forcasting process more efficient. The offline precomputation stage is conducted before the inference, where we define a full-computation schedule based on the schedule parameters and pre-compute the interpolation coefficients for the steps that are not in the full-computation schedule. The online application stage is conducted during the inference, where we initialize an empty history buffer to store the features computed at full-computation steps, and use the pre-computed coefficients to approximate the features at the extrapolation steps.

\begin{algorithm}[H]
\caption{ChebBooster Algorithm}
\label{alg:chebbooster_pseudocode}
\begin{algorithmic}[1]
\Statex \textbf{Require:} Model \(M\), schedule parameters \((s_0, s_1, r)\), history size \(n\).

\Statex \textbf{1. Offline Precomputation Stage:}
\State Define full-computation schedule \(\mathcal{F}\) based on \(s_0, s_1, r\).
\State For each step \(t \notin \mathcal{F}\), pre-compute and store interpolation coefficients \(\boldsymbol{w}^t\).

\Statex \textbf{2. Online Application Stage:}
\State Initialize an empty history buffer \(H\) (size \(n\)).
\For{each timestep \(s\) from \(0\) to \(T-1\)}
    \If{\(s \in \mathcal{F}\)} \Comment{Full computation}
        \State Compute feature \(f_s\) using the model \(M\).
        \State Store the pair \((s, f_s)\) in the history buffer \(H\).
    \Else \Comment{Extrapolation}
        \State Retrieve cached features \(\{f_j\}\) from \(H\).
        \State Retrieve pre-computed coefficients \(\boldsymbol{w^s}\).
        \State Approximate feature \(f_s \gets \sum \alpha^s_j \cdot f_j\).
    \EndIf
    \State Use the feature \(f_s\) to proceed with the diffusion step.
\EndFor
\end{algorithmic}
\end{algorithm}

\section{Additional Introduction to Experiment Settings}

Our experiments on DiT-XL/2 and PixArt-\(\Sigma\) are conducted on Nvidia RTX 4090 GPUs, while the experiments on FLUX.1-dev are conducted on an Nvidia A800 GPU. For the experiments of PixArt-\(\Sigma\), we seperate the whole process into the text embedding process and the sampling process, for the former is conducted redundantly with the same prompts. This step allows us to conduct all of our experiments on only one low-memory GPU. The ImageReward score is evaluated using their official evaluation code, and the evaluation model is ImageReward=1.0, which can be retrieved from the Hugging Face model hub. The ClipScore is evaluated using the code implemented by torchmetrics, and the evaluation model is clip-vit-base-patch32, which can also be retrieved from the Hugging Face model hub. The evaluation code of FID, sFID and Inception Score is from the official implementation of Guided Diffusion. In the C2I task, we set the CFG scale to 1.55, and set the seed to 2025. For the T2I tasks, we set the seed to 2025 and keep the same settings as the original implementation.

Here we notice there is a mistake in the Reproducibility Checklist, for the Question 4.6, our answer should be "Yes". We apologize for the missing of the answer.

\section{Additional Experiments on DiT-XL/2}

\newcommand{\tableCompression}{0.85}

\begin{table}[ht]

    {
    \renewcommand{\arraystretch}{\tableCompression}
    \centering

    \small
    \setlength{\tabcolsep}{5pt}
    
    \begin{tabular}{l|ccc|c}
    \toprule
    \textbf{Method} & {\textbf{IS$\uparrow$}} & {\textbf{FID$\downarrow$}} & {\textbf{sFID$\downarrow$}} & {\textbf{FLOPs$\downarrow$}} \\
    \midrule
    
    Original ($50$ steps) & 251.56 & 2.18 & 4.29 & 23.735 \\
    Original ($25$ steps) & 244.03 & 2.75 & 4.55 & 11.868 \\
    Original ($20$ steps) & 235.35 & 3.27 & 4.93 & 9.494 \\
    Original ($16$ steps) & 223.38 & 4.24 & 5.69 & 7.595 \\
    Original ($12$ steps) & 195.83 & 7.07 & 8.05 & 5.696 \\
    Original ($10$ steps) & 168.99 & 11.19 & 11.12 & 4.747 \\
    \midrule
    
    Cheb ($r=2, n=2$) & 246.98 & 2.40 & 4.91 & 8.562 \\
    Cheb ($r=3, n=3$) & 246.49 & 2.25 & 4.59 & 8.564 \\
    Cheb ($r=4, n=4$) & 247.27 & 2.40 & 4.84 & 8.566 \\
    Cheb ($r=5, n=5$) & 247.27 & 2.28 & 4.64 & 8.567 \\
    \midrule
    
    Cheb ($r=2, n=2$) & 244.54 & 2.62 & 5.63 & 6.666 \\
    Cheb ($r=3, n=3$) & 242.58 & 2.35 & 4.95 & 6.668 \\
    Cheb ($r=4, n=4$) & 244.31 & 2.61 & 5.44 & 6.670 \\
    Cheb ($r=5, n=5$) & 244.20 & 2.38 & 5.02 & 6.671 \\
    \midrule
    
    Cheb ($r=2, n=2$) & 240.99 & 2.79 & 5.67 & 5.717 \\
    Cheb ($r=3, n=3$) & 238.48 & 2.44 & 4.96 & 5.720 \\
    Cheb ($r=4, n=4$) & 241.06 & 2.75 & 5.46 & 5.721 \\
    Cheb ($r=5, n=5$) & 241.26 & 2.49 & 4.97 & 5.722 \\
    \midrule
    
    Cheb ($r=2, n=2$) & 230.61 & 3.50 & 7.68 & 4.769 \\
    Cheb ($r=3, n=3$) & 228.95 & 2.87 & 5.76 & 4.772 \\
    Cheb ($r=4, n=4$) & 230.38 & 3.33 & 6.78 & 4.773 \\
    Cheb ($r=5, n=5$) & 232.37 & 2.91 & 5.85 & 4.774 \\
    \midrule
    
    Cheb ($r=2, n=2$) & 223.52 & 3.98 & 8.32 & 4.295 \\
    Cheb ($r=3, n=3$) & 223.52 & 3.17 & 5.91 & 4.297 \\
    Cheb ($r=4, n=4$) & 226.86 & 3.60 & 6.94 & 4.299 \\
    Cheb ($r=5, n=5$) & 228.57 & 3.14 & 5.68 & 4.299 \\
    \bottomrule
    \end{tabular}
    }
\caption{\textbf{Performance Comparison of ChebBooster on DiT-XL/2.} This table summarizes the performance of ChebBooster across different configurations, comparing Inception Score (IS), FID, sFID, and FLOPs against the original model with varying sampling steps. The results demonstrate that ChebBooster achieves significant speedup while maintaining competitive quality metrics.}
\label{tab:performance_comparison_final}
\end{table}

\begin{figure}[h]
    \centering
    \includegraphics[width=\linewidth]{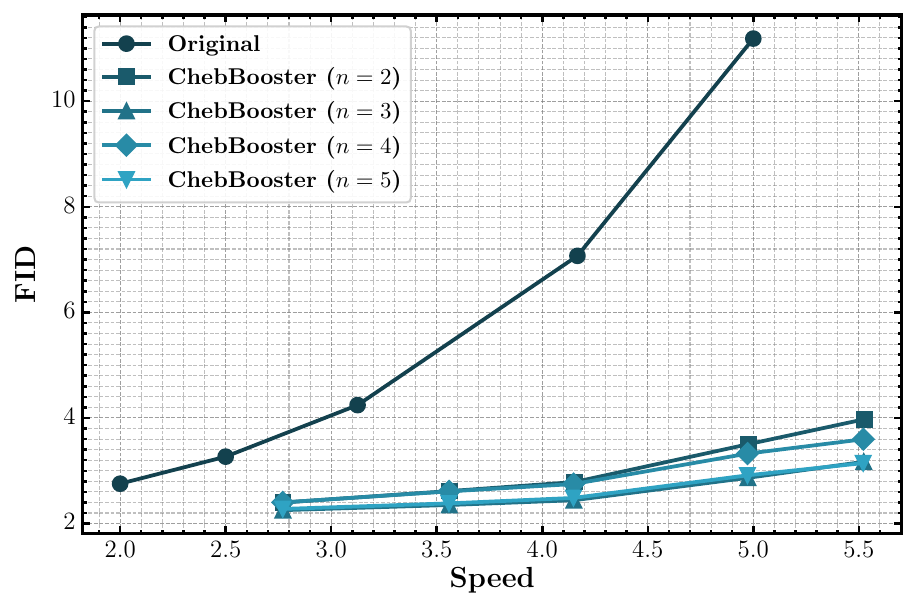}
    \caption{\textbf{FID evaluation on DiT-XL/2 across varying sampling steps.} This figure shows the trade-off between FID scores and inference speed when applying ChebBooster. Notably, ChebBooster achieves substantial acceleration while preserving competitive FID performance, indicating that the generated images remain close to the real data distribution even under aggressive speedup.}
    \label{fig:ditxl2_additional_experiments_fid}
\end{figure}

\begin{figure}[h]
    \centering
    \includegraphics[width=\linewidth]{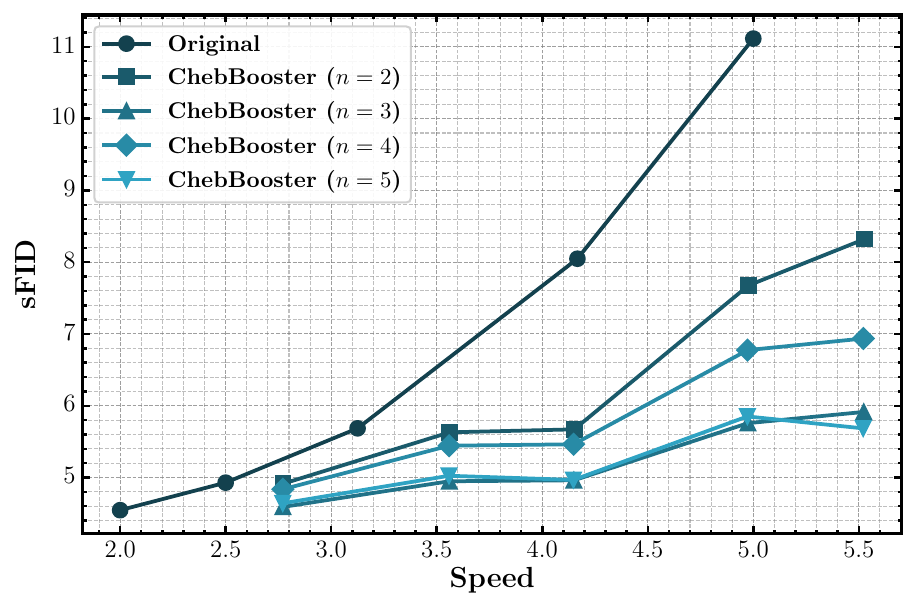}
    \caption{\textbf{Structural FID (sFID) evaluation on DiT-XL/2.} The figure highlights how ChebBooster maintains structural consistency in generated images while achieving high speedup ratios. Compared to FID, sFID is more sensitive to localized distortions, and the consistently low sFID values indicate that structural fidelity is preserved even at faster sampling rates.}
    \label{fig:ditxl2_additional_experiments_sfid}
\end{figure}

\begin{figure}[h]
    \centering
    \includegraphics[width=\linewidth]{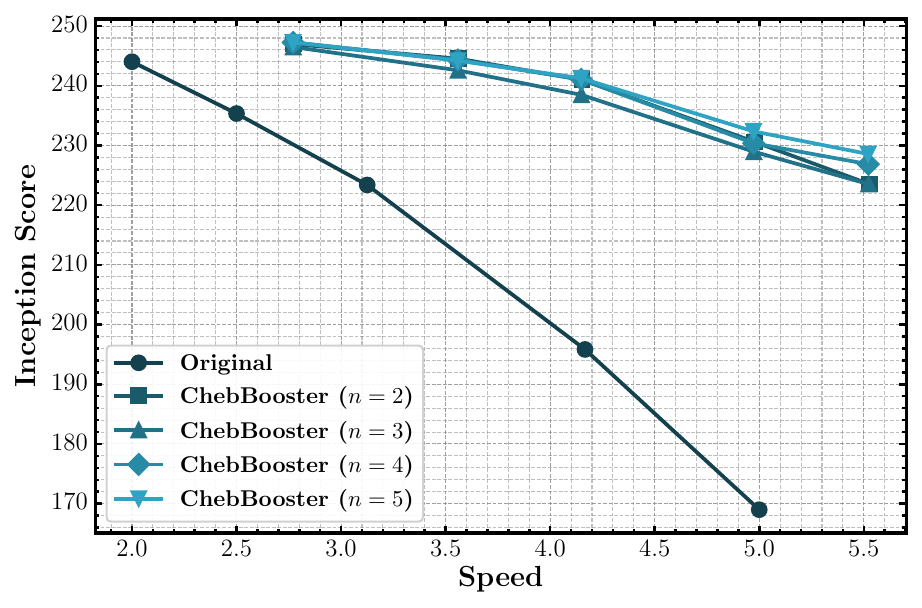}
    \caption{\textbf{Inception Score evaluation on DiT-XL/2.} This figure shows how ChebBooster impacts sample diversity and semantic clarity across different acceleration levels. Despite fewer sampling steps, the method sustains high Inception Scores, indicating that the generated samples remain both varied and classifiable.}
    \label{fig:ditxl2_additional_experiments_inception_score}
\end{figure}

\begin{figure*}
    \centering
    \includegraphics[width=0.75\linewidth]{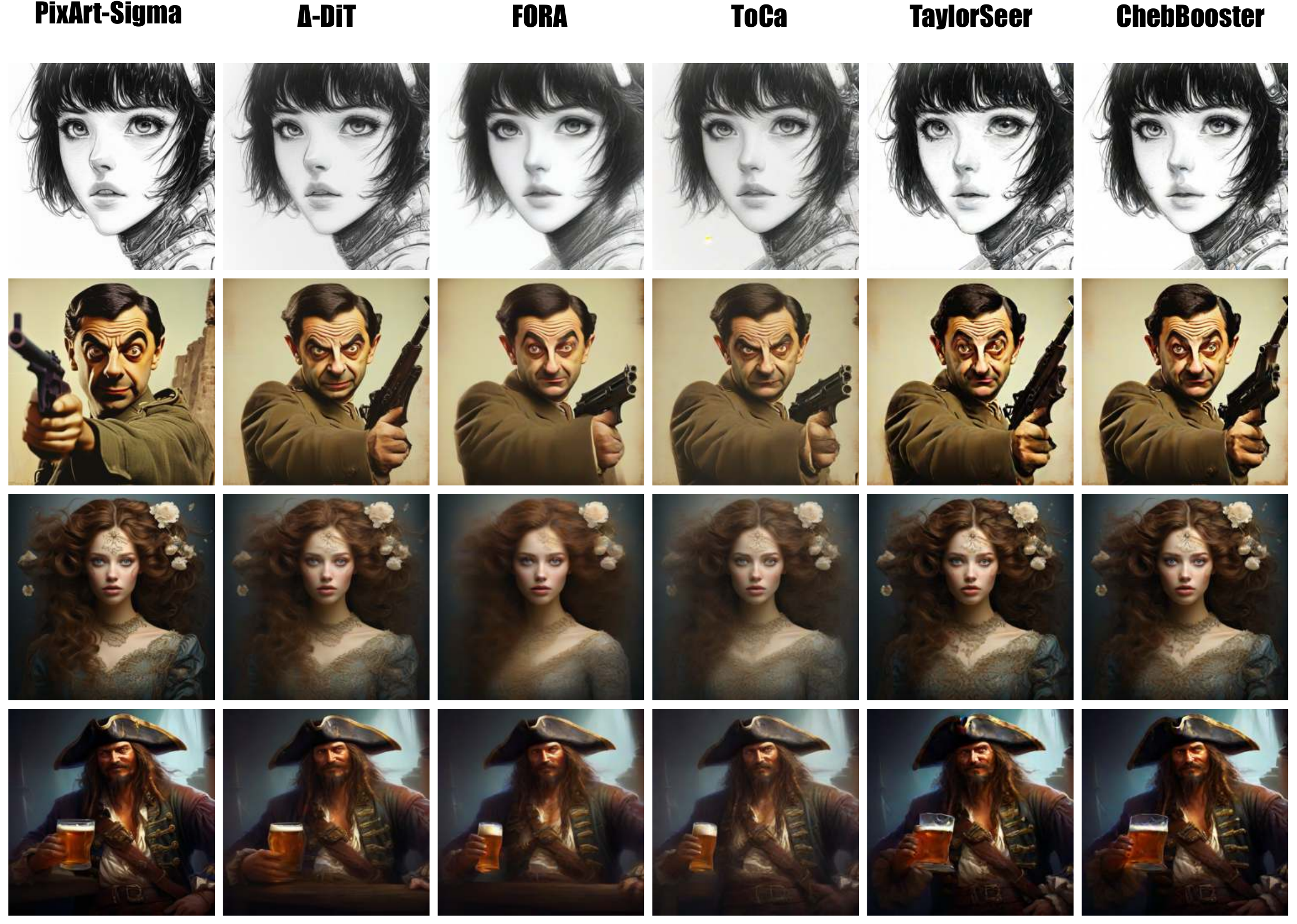}
    \caption{\centering \textbf{Extensive qualitative results on PixArt-\(\Sigma\).}}
    \label{fig:pixart_sigma_prompts}
\end{figure*}

\begin{figure*}
    \centering
    \includegraphics[width=0.93\linewidth]{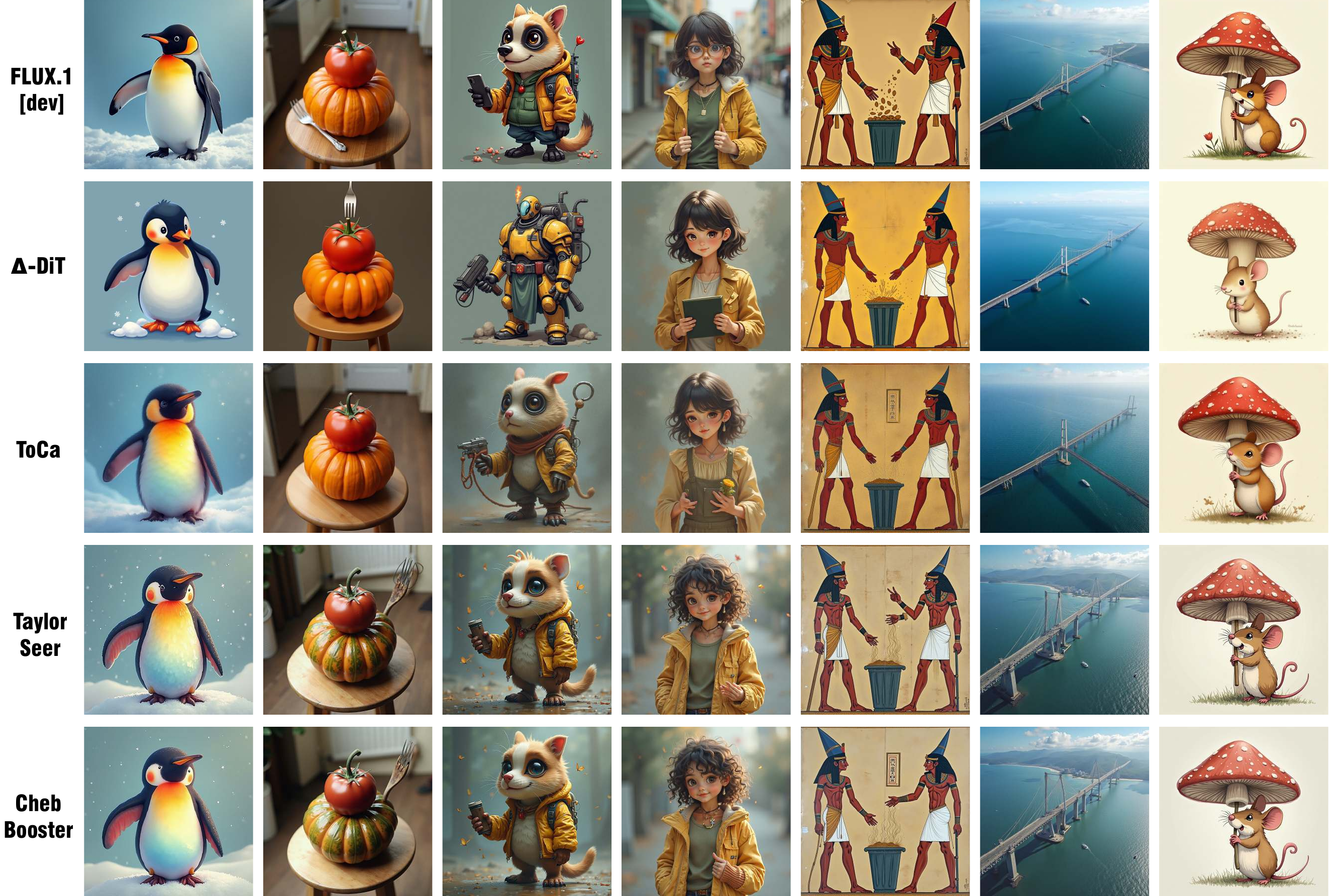}
    \caption{\centering \textbf{Extensive qualitative results on FLUX.1-dev.}}
    \label{fig:flux_prompts}
\end{figure*}

We conduct additional experiments on DiT-XL/2 to further validate the effectiveness of ChebBooster. The results are shown in Table~\ref{tab:performance_comparison_final}, and the visualization has been illustrated in Figure~\ref{fig:ditxl2_additional_experiments_fid}, Figure~\ref{fig:ditxl2_additional_experiments_sfid} and Figure~\ref{fig:ditxl2_additional_experiments_inception_score}. Our evaluation focuses on comparing the proposed ChebBooster method against the Original baseline, performing an ablation study on the hyperparameters of ChebBooster, and offering optimal parameter recommendations based on the empirical results.

\noindent \textbf{Key Observations.}
Our experiments reveal several critical insights. First, the Original models exhibit a clear and predictable trade-off: reducing inference steps (e.g., from 50 to 10) cuts FLOPs by nearly 80\% but causes a catastrophic degradation in generation quality, with the FID score worsening by over 410\% (from 2.18 to 11.19). This establishes a critical performance bottleneck for naive acceleration. Second, ChebBooster consistently overcomes this limitation. For instance, \texttt{Cheb (r=5, n=5)} achieves a superior FID (2.28 vs. 3.27) and a much higher IS (247.27 vs. 235.35) than \texttt{Original (20 steps)} while requiring \textbf{10\% fewer FLOPs}. This demonstrates that ChebBooster fundamentally improves the performance-efficiency frontier. Finally, the hyperparameters have distinct roles: $r$ primarily governs the foundational acceleration and computational budget, while $n$ acts as a quality-tuning parameter within that budget, consistently improving fidelity for a negligible computational cost.

\noindent \textbf{Parameter Recommendations.}
Based on this analysis, we can recommend specific parameter configurations tailored to different use-cases, depending on the desired balance between generation fidelity and computational efficiency.

\noindent \textbf{For Maximizing Generation Quality.}
If the primary goal is to achieve the best possible output quality while still realizing significant speedup, the optimal choice is \textbf{\texttt{Cheb (r=3, n=5)}}. It yields the highest Inception Score (247.27) and one of the best FID scores (2.28) among all accelerated models. Its performance is nearly on par with the \texttt{Original (50 steps)} baseline (FID of 2.28 vs. 2.18) but reduces the computational cost by over 63\%.

\noindent \textbf{For Maximizing Computational Efficiency.}
If the primary constraint is minimizing FLOPs for deployment on resource-limited hardware, the recommended configuration is \textbf{\texttt{Cheb (r=7, n=2)}}. It offers the lowest computational footprint (4.295 GFLOPs) of all tested ChebBooster variants. Critically, when compared to the \texttt{Original (10 steps)} model which has a similar computational budget, this \texttt{ChebBooster} configuration is vastly superior, improving the FID from a poor 11.19 down to a respectable 3.98.

\noindent \textbf{A Balanced Recommendation.}
For a general-purpose, high-performance setting, a configuration of \textbf{$r=4$} with \textbf{$n \ge 4$} provides an excellent balance. For example, \texttt{Cheb (r=4, n=5)} maintains a strong FID of 2.38 and an IS of 244.20, while reducing FLOPs to 6.671---a 72\% reduction from the full model.

In summary, ChebBooster is a robust and effective method for accelerating the generative model. By appropriately selecting the acceleration level $r$ and tuning the quality with $n$, users can achieve performance far superior to naive methods across the entire performance-efficiency spectrum.

\section{Extensive Visual Quality Analysis of ChebBooster}

To further validate the effectiveness of ChebBooster, we conducted visual comparisons on both class-to-image and text-to-image generation tasks, as shown in Figure~\ref{fig:pixart_sigma_prompts} and Figure~\ref{fig:flux_prompts}. These comparisons include several recent training-free acceleration methods---$\Delta$-DiT, FORA, ToCa, and TaylorSeer---as baselines. Across all evaluated prompts and classes, ChebBooster consistently produces outputs that are visually superior in terms of sharpness, semantic fidelity, and structural coherence.

In the PixArt-Sigma setting (Figure~\ref{fig:pixart_sigma_prompts}), ChebBooster avoids the blurriness and semantic degradation exhibited by TaylorSeer and $\Delta$-DiT. While ToCa and FORA generate outputs with acceptable global layout, they often suffer from washed-out textures or over-simplified details. In contrast, ChebBooster maintains fine-grained textures, well-defined contours, and vivid colors, demonstrating its ability to preserve both high-frequency details and global consistency.

Similarly, in the FLUX.1-dev setting (Figure~\ref{fig:flux_prompts}), ChebBooster clearly outperforms all other baselines in generating semantically correct and visually appealing images. Notably, ChebBooster retains distinctive visual attributes (e.g., facial features, object poses, and stylistic elements) that are often lost or distorted in other methods. This indicates that the Chebyshev-inspired extrapolation strategy employed by ChebBooster not only improves sampling efficiency, but also enhances the model's ability to maintain feature consistency across denoising steps.

Overall, these qualitative results provide strong evidence that ChebBooster achieves a better trade-off between acceleration and image quality. Its robustness across different models and prompts further suggests that ChebBooster generalizes well to diverse generative scenarios.

\section{Prompts for Demonstration}

We provide the prompts used in our demonstration of ChebBooster on PixArt-\(\Sigma\) and FLUX.1-dev in Figure~\ref{fig:pixart_comparison} and 6. These prompts are designed to showcase the capabilities of ChebBooster in generating high-quality images with various styles and subjects.

\begin{enumerate}[leftmargin=*, label=\arabic*.]
    \item An Ultraman preparing to take flight. (Figure~\ref{fig:pixart_comparison})
    \item A full body shot of an elegant, Scottish woman wearing a dress with a sharp focus on her striking eyes in a realistic and beautifully retouched art piece by Artgerm and Jason Chan. (Figure~\ref{fig:pixart_comparison})
    \item A portrait painting of a red-haired, smiling woman in a green dress against a golden background with intricate patterns. (Figure~\ref{fig:pixart_comparison})
    \item Yoshitaka Amano's painting of a young lion beastman with a white mane, wearing complex fantasy clothing and huge paws, at a medieval market on a windy day. (Figure~\ref{fig:pixart_comparison})
    \item A white toilet sitting next to a large window. (mistaken for "A white toilet tin a bathroom sitting next to a sink." in Figure~\ref{fig:pixart_comparison})
    \item Two people standing in a kitchen near a stove. (Figure~\ref{fig:pixart_comparison})
    \item A single clock is sitting on a table. (Figure~\ref{fig:comparison_result_flux})
    \item An illustration of a small green elephant standing behind a large red mouse. (Figure~\ref{fig:comparison_result_flux})
    \item Medium shot black and white manga pencil drawing with a highly detailed face of Alita by Yukito Kishiro. (Figure~\ref{fig:comparison_result_flux})
    \item Mr. Bean featured on a WWII propaganda poster holding a gun. (Figure~\ref{fig:pixart_sigma_prompts})
    \item The image portrays Ophelia with a detailed and elegant face, featuring wonderful eyes, wearing an intricate dress, and created with hyperrealistic painting techniques. (Figure~\ref{fig:pixart_sigma_prompts})
    \item A pirate with a beer is illustrated in detailed digital painting. (Figure~\ref{fig:pixart_sigma_prompts})
    \item Rainbow coloured penguin. (Figure~\ref{fig:flux_prompts})
    \item A tomato has been put on top of a pumpkin on a kitchen stool. There is a fork sticking into the pumpkin. The scene is viewed from above. (Figure~\ref{fig:flux_prompts})
    \item Rbefraigerator. (Figure~\ref{fig:flux_prompts})
    \item Matutinal. (Figure~\ref{fig:flux_prompts})
    \item An ancient Egyptian painting depicting an argument over whose turn it is to take out the trash. (Figure~\ref{fig:flux_prompts})
    \item A bridge connecting Europe and North America on the Atlantic Ocean, bird's eye view. (Figure~\ref{fig:flux_prompts})
    \item Illustration of a mouse using a mushroom as an umbrella. (Figure~\ref{fig:flux_prompts})
\end{enumerate}

\vspace{1em}

\end{document}